\documentclass{article}
\usepackage[final]{corl_2018}
\usepackage[usenames,dvipsnames]{xcolor}

\usepackage{multicol}
\usepackage{array}
\usepackage{booktabs}
\usepackage{multirow}
\usepackage{amsfonts}
\usepackage{enumitem}
\usepackage{soul}

\usepackage{graphics} 
\usepackage{epsfig} 
\usepackage{times} 
\usepackage{amsmath} 
\usepackage{tabularx}
\usepackage{graphicx} 
\usepackage{microtype}
\usepackage{multirow}
\usepackage{algorithmic}
\usepackage[ruled,linesnumbered,vlined]{algorithm2e}
\usepackage[keeplastbox]{flushend}
\usepackage{comment}

\title{Expanding Motor Skills using Relay Networks}

\author{
  Visak CV Kumar\\
  Georgia Institute of Technology\\
  \texttt{visak3@gatech.edu} \\
   \And
    Sehoon Ha \\
    Google Brain\\
     \texttt{sehoonha@google.com} \\
   \And
     C.Karen Liu \\
     Georgia Institute of Technology \\
     \texttt{karenliu@cc.gatech.edu} \\
}

\newcommand{\cmt}[1]{}

\newcommand{\karen}[1]{\textcolor{red}{{Karen: #1}}}

\newcommand{\sehoon}[1]{\textcolor{magenta}{{Sehoon: #1}}}
\newcommand{\sehoonedit}[1]{\textcolor{magenta}{{#1}}}

\long\def\ignorethis#1{}

\newcommand{\etal}{{et~al.\ }}
\newcommand{\eg}{e.g.\ }
\newcommand{\ie}{i.e.\ }

\newcommand{\figref}[1]{Fig.~\ref{fig:#1}}
\renewcommand{\eqref}[1]{Eq. ~(\ref{eq:#1})}
\newcommand{\secref}[1]{Section~\ref{sec:#1}}
\newcommand{\tabref}[1]{Table~\ref{tab:#1}}
\newcommand{\algref}[1]{Algorithm~\ref{alg:#1}}

\newcommand{\vc}[1]{\ensuremath{\mathbf{#1}}}

\newcommand{\argmin}{\operatornamewithlimits{argmin}}

\newcommand{\pctab}{\hspace{0.2in}}

\begin{document}

\maketitle


\begin{abstract}
While recent advances in deep reinforcement learning have achieved impressive results in learning motor skills, many policies are only capable within a limited set of initial states. We propose an algorithm that sequentially decomposes a complex robotic task into simpler subtasks and trains a local policy for each subtask such that the robot can expand its existing skill set gradually. Our key idea is to build a directed graph of local control policies represented by neural networks, which we refer to as \emph{relay neural networks}. Starting from the first policy that attempts to achieve the task from a small set of initial states, the algorithm iteratively discovers the next subtask with increasingly more difficult initial states until the last subtask matches the initial state distribution of the original task. The policy of each subtask aims to drive the robot to a state where the policy of its preceding subtask is able to handle. By taking advantage of many existing actor-critic style policy search algorithms, we utilize the optimized value function to define ``good states'' for the next policy to relay to. 
\end{abstract}


\section{Introduction}
	
The recent advances in deep reinforcement learning have motivated robotic researchers to tackle increasingly difficult control problems. For example, OpenAI RoboSchool \cite{roboschool} challenges the researchers to create robust control policies capable of locomotion while following high-level goals under external perturbation. One obvious approach is to use a powerful learning algorithm and a large amount of computation resources to learn a wide range of situations. While this direct approach might work occasionally, it is difficult to scale up with the ever-increasing challenges in robotics.

Alternatively, we can consider to break down a complex task into a sequence of simpler subtasks and solve each subtask sequentially. The control policy for each subtask can be trained, sequentially as well, to expand existing skill set gradually. While an arbitrary decomposition of a task might lead to suboptimal control policy, a robot performing actions sequentially is often considered a practical strategy to complete a complex task reliably \cite{vukobratovic1970stability,stuckler2006getting}, albeit not optimally. 

This paper introduces a new technique to expand motor skills by connecting new policies to existing ones. In the similar spirit of hierarchical reinforcement learning (HRL) \cite{Dietterich:2000:HRL,Barto:2003:RAH} which decomposes a large-scale Markov Decision Process (MDP) into subtasks, the key idea of this work is to build a directed graph of local policies represented by neural networks, which we refer to as \emph{relay neural networks}. Starting from the first policy that attempts to achieve the task from a small set of initial states, the algorithm gradually expands the set of successful initial states by connecting new policies, one at a time, to the existing graph until the robot can achieve the original complex task. Each policy is trained for a new set of initial states with an objective function that encourages the policy to drive the new states to existing ``good'' states. 

The main challenge of this method is to determine the ``good'' states from which following the relay networks will eventually achieve the task. Fortunately, a few modern policy learning algorithms, such as TRPO-GAE \cite{schulman2015high}, PPO \cite{schulman2017proximal}, DDPG \cite{lillicrap2015continuous}, or A3C \cite{mnih2016asynchronous}, provide an estimation of the value function along with the learned policy. Our algorithm utilizes the value function to train a policy that relays to its preceding policy, as well as to transition the policies on the graph during online execution. We demonstrate that the relay networks can solve complex control problems for underactuated systems.
\section{Related Works}

\textbf{Hierarchical Reinforcement Learning:}
HRL has been proven successful in learning policies for large-scale problems. This approach decomposes problems using \emph{temporal abstraction} which views sub-policies as macro actions, or \emph{state abstraction} which focuses on certain aspects of state variables relevant to the task. Prior work \cite{sutton1999between,daniel2012hierarchical,kulkarni2016hierarchical,heess2016learning,peng2017deeploco} that utilizes temporal abstraction applies the idea of parameterized goals and pseudo-rewards to train macro actions, as well as training a meta-controller to produce macro actions. One notable work that utilizes state abstraction is MAXQ value function decomposition which decomposes a large-scale MDP into a hierarchy of smaller MDP's \cite{Dietterich:2000:HRL,Bai:2012:OPL,Grave:2014:BEI}. MAXQ enables individual MDP's to only focus on a subset of state space, leading to better performing policies. Our relay networks can be viewed as a simple case of MAXQ in which the recursive subtasks, once invoked, will directly take the agent to the goal state of the original MDP. That is, in the case of relay networks, the Completion Function that computes the cumulative reward after the subtask is finished always returns zero. As such, our method avoids the need to represent or approximate the Completion Function, leading to an easier RL problem for continuous state and action spaces.

\textbf{Chaining and Scheduling Control Policies:}
Many researchers \cite{burridge1999sequential} have investigated the idea of chaining a new policy to existing policies.
Tedrake \cite{tedrake2009lqr} proposed the LQR-Tree algorithm that combines locally valid linear quadratic regulator (LQR) controllers into a nonlinear feedback policy to cover a wider region of stability. 
Borno \etal \cite{borno2017domain} further improved the sampling efficiency of RRT trees \cite{tedrake2009lqr} by expanding the trees with progressively larger subsets of initial states. However, the success metric for the controllers is based on Euclidean distance in the state space, which can be inaccurate for high dimensional control problems with discontinuous dynamics. Konidaris \etal \cite{konidaris2009skill} proposed to train a chain of controllers with different initial state distributions. The rollouts terminate when they are sufficiently close to the initial states of the parent controller. They discovered an initiation set using a sampling method in a low dimensional state space. In contrast, our algorithm utilizes the value function of the parent controller to modify the reward function and define the terminal condition for policy training. There also exists a large body of work on scheduling existing controllers, such as controllers designed for taking simple steps \cite{coros2009robust} or tracking short trajectories \cite{liu2017learning}. Inspired by the prior art, our work demonstrates that the idea of sequencing a set of local controllers can be realized by learning policies represented by the neural networks.

\textbf{Learning with Structured Initial States:}
Manipulating the initial state distribution during training has been considered a promising approach to accelerate the learning process. \citet{kakade2002approximately} studied theoretical foundation of using  ``exploratory'' restart distribution for improving the policy training.
Recently, \citet{popov2017data} demonstrated that the learning can be accelerated by taking the initial states from successful trajectories.
\citet{florensa17a} proposed a reverse curriculum learning algorithm for training a policy to reach a goal using a sequence of initial state distributions increasingly further away from the goal. We also train a policy with reversely-ordered initial states, but we exploited chaining mechanism of multiple controllers rather than training a single policy.

\textbf{Learning Policy and Value Function:}
Our work builds on the idea of using value functions to provide information for connecting policies. The approach of actor-critic \cite{sutton2000policy,konda2000actor} explicitly optimizes both the policy (actor) and the value function (critic) such that the policy update direction can be computed with the help of critic. Although the critic may introduce bias, the policy update can be less subject to variance compared to pure policy gradient methods, such as REINFORCE \cite{williams1992simple}. Recently, \citet{lillicrap2015continuous} proposed Deep Deterministic Policy Gradient (DDPG) that combines the actor-critic approach \cite{silver2014deterministic} with Deep Q Network (DQN) \cite{mnih2015human} for solving high-dimensional continuous control problems. Schulman \etal \cite{schulman2015high} proposed the general advantage estimator (GAE) for reducing the variance of the policy gradient while keeping the bias at a reasonable level. Further, Mnih \etal \cite{mnih2016asynchronous} demonstrated a fast and robust learning algorithm for actor-critic networks using an asynchronous update scheme. Using these new policy learning methods, researchers have demonstrated that policies and value functions represented by neural networks can be applied to high-dimensional complex control problems with continuous actions and states \cite{peng2016terrain,peng2017deeploco,kumar2017learning}. Our algorithm can work with any of these policy optimization methods, as long as they provide a value function approximately consistent with the policy.

\section{Method}
Our approach to a complex robotic task is to automatically decompose it to a sequence of subtasks, each of which aims to reach a state where the policy of its preceding subtask is able to handle. The original complex task is formulated as a MDP described by a tuple $\{\mathcal{S}, \mathcal{A}, r, \mathcal{T}, \rho, P\}$, where $\mathcal{S}$ is the state space, $\mathcal{A}$ is the action space, $r$ is the reward function, $\mathcal{T}$ is the set of termination conditions, $\rho=N(\boldsymbol{\mu}_\rho, \Sigma_\rho)$ is the initial state distribution, and $P$ is the transition probability. Instead of solving for a single policy for the MDP, our algorithm solves for a set of policies and value functions for a sequence of simpler MDP's. A policy, $\pi: \mathcal{S} \times \mathcal{A} \mapsto [0, 1]$, is represented as a Gaussian distribution of action $\vc{a} \in \mathcal{A}$ conditioned on a state $\vc{s} \in \mathcal{S}$. The mean of the distribution is represented by a fully connected neural network and the covariance is defined as part of the policy parameters to be optimized. A value function, $V: \mathcal{S} \mapsto R$, is also represented as a neural network whose parameters are optimized by the policy learning algorithm. 

We organize the MDP's and their solutions in a directed graph $\Gamma$. A node $\mathcal{N}_k$ in $\Gamma$ stores the initial state distribution $\rho_k$ of the $k^{th}$ MDP, while a directed edge connects an MDP to its parent MDP and stores the solved policy ($\pi_k$), the value function ($V_k$), and the threshold of value function ($\bar{V}_k$) (details in Section \ref{sec:threshold}). As the robot expands its skill set to accomplish the original task, a chain of MDP's and policies is developed in $\Gamma$. If desired, our algorithm can be extended to explore multiple solutions to achieve the original task. Section \ref{sec:multiple-chains} describes how multiple chains can be discovered and merged in $\Gamma$ to solve multi-modal problems.

\subsection{Learning Relay Networks}
\label{sec:newnodes}
The process of learning the relay networks (Algorithm \ref{alg:relay_networks}) begins with defining a new initial state distribution $\rho_0$ which reduces the difficulty of the original MDP (Line \ref{line:initial}). Although our algorithm requires the user to define $\rho_0$, it is typically intuitive to find a $\rho_0$ which leads to a simpler MDP. For example, we can define $\rho_0$ as a Gaussian whose mean, $\boldsymbol{\mu}_{\rho_0}$, is near the goal state of the problem.

Once $\rho_0$ is defined, we proceed to solve the first subtask $\{\mathcal{S}, \mathcal{A}, r, \mathcal{T}, \rho_0, P\}$, whose objective function is defined as the expected accumulated discounted rewards along a trajectory:
\begin{equation}
\label{eqn:rootObj}
J_0 = \mathbb{E}_{\vc{s}_{0:t_f}, \vc{a}_{0:t_f}}[\sum_{t=0}^{t_f} \gamma^t r(\vc{s}_t, \vc{a}_t)],
\end{equation}
where $\gamma$ is the discount factor, $\vc{s}_0$ is the initial state of the trajectory drawn from $\rho_0$, and $t_f$ is the terminal time step of the trajectory. We can solve the MDP using PPO/TRPO or A3C/DDPG to obtain the policy $\pi_0$, which drives the robot from a small set of initial states from $\rho_0$ to states where the original task is completed, as well as the value function $V_0(\vc{s})$, which evaluates the return by following $\pi_0$ from state $\vc{s}$ (Line \ref{line:solve_0}).

\begin{algorithm}[t]
\caption{LearnRelayNetworks}\label{alg:relay_networks}
\begin{algorithmic}[1]
\STATE \textbf{Input:} MDP $\{\mathcal{S}, \mathcal{A}, r, \mathcal{T}, \rho, P\}$
\STATE Add root node, $\bar{\mathcal{N}} = \{\emptyset\}$, to $\Gamma$
\STATE Define a simpler initial state distribution $\rho_0$ \label{line:initial} 
\STATE Define objective function $J_0$ according to Equation \ref{eqn:rootObj}
\STATE $[\pi_0, V_0] \leftarrow$ \textrm{PolicySearch}($\mathcal{S}, \mathcal{A}, \rho_0, J_0, \mathcal{T})$ \label{line:solve_0}
\STATE $\bar{V}_0 \leftarrow$
ComputeThreshold($\pi_0, V_0, \mathcal{T}, \rho_0, J_0$) \label{line:threshold_0}
\STATE Add node, $\mathcal{N}_0 = \{\rho_0\}$, to $\Gamma$ \label{line:addnode_0}
\STATE Add edge, $\mathcal{E} = \{\pi_0, V_0, \bar{V}_0, \}$, from $\mathcal{N}_0$ to $\bar{\mathcal{N}}$\label{line:addedge_0}
\STATE $k=0$
\WHILE{$\pi_k$ does not succeed from $\rho$} \label{line:while}
\STATE $\mathcal{T}_{k+1} = \mathcal{T} \cup (V_{k}(\vc{s}) > \bar{V}_{k})$ \label{line:terminal}
\STATE $\rho_{k+1} \leftarrow$ Compute new initial state distribution using Equation \ref{eqn:discoverNewNode} \label{line:next_initial}
\STATE Define objective function $J_{k+1}$ according to Equation \ref{eqn:relay_obj}
\STATE $[\pi_{k+1}, V_{k+1}] \leftarrow$ \textrm{PolicySearch}($\mathcal{S}, \mathcal{A}, \rho_{k+1}, J_{k+1}, \mathcal{T}_{k+1})$ \label{line:solve}
\STATE $\bar{V}_{k+1} \leftarrow$
ComputeThreshold($\pi_{k+1}, V_{k+1}, \mathcal{T}, \rho_{k+1}, J_{k+1}$) \label{line:threshold}
\STATE Add node, $\mathcal{N}^i_{k+1} = \{\rho_{k+1}\}$, to $\Gamma$ \label{line:addnode}
\STATE Add edge, $\mathcal{E} = \{\pi_{k+1}, V_{k+1}, \bar{V}_{k+1}\}$, from node $\mathcal{N}_{k+1}$ to $\mathcal{N}_k$ \label{line:addedge}
\STATE $k \leftarrow k+1$
\ENDWHILE
\RETURN{$\Gamma$}
\end{algorithmic}
\end{algorithm}

The policy for the subsequent MDP aims to drive the rollouts toward the states which $\pi_0$ can successfully handle, instead of solving for a policy that directly completes the original task. To determine whether $\pi_0$ can handle a given state $\vc{s}$, one can generate a rollout by following $\pi_0$ from $\vc{s}$ and calculate its return. However, this approach can be too slow for online applications. Fortunately, many of the modern policy gradient methods, such as PPO or A3C, produce a value function, which provides an approximated return from $\vc{s}$ without generating a rollout. Our goal is then to determine a threshold $\bar{V}_0$ for $V_0(\vc{s})$ above which $\vc{s}$ is deemed ``good'' (Line \ref{line:threshold_0}). The details on how to determine such a threshold are described in Section \ref{sec:threshold}. We can now create the first  node $\mathcal{N}_0 = \{\rho_0\}$ and add it to $\Gamma$, as well as an edge $\mathcal{E} = \{\pi_0, V_0, \bar{V}_0\}$ that connects $\mathcal{N}_0$ to the dummy root node $\bar{\mathcal{N}}$ (Line \ref{line:addnode_0}-\ref{line:addedge_0}).

Starting from $\mathcal{N}_0$, the main loop of the algorithm iteratively adds more nodes to $\Gamma$ by solving subsequent MDP's until the last policy $\pi_k$, via relaying to previous policies $\pi_{k-1}, \cdots \pi_0$, can generate successful rollouts from the states drawn from the original initial state distribution $\rho$ (Line \ref{line:while}). At each iteration $k$, we formulate the $(k+1)^{th}$ subsequent MDP by redefining $\mathcal{T}_{k+1}$, $\rho_{k+1}$, and the objective function $J_{k+1}$, while using the shared $\mathcal{S}$, $\mathcal{A}$, and $P$. First, we define the terminal conditions $\mathcal{T}_{k+1}$ as the original set of termination conditions ($\mathcal{T}$) augmented with another condition, $V_{k}(\vc{s}) > \bar{V}_{k}$ (Line \ref{line:terminal}). Next, unlike $\rho_0$, we define the initial state distribution $\rho_{k+1}$ through an optimization process. The goal of the optimization is to find the mean of the next initial state distribution, $\boldsymbol{\mu}_{\rho_{k+1}}$, that leads to unsuccessful rollouts under the current policy $\pi_k$, without making the next MDP too difficult to relay. To balance this trade-off, the optimization moves in a direction that reduces the value function of the most recently solved MDP, $V_k$, until it reaches the boundary of the ``good state zone'' defined by $V_k(\vc{s})\geq \bar{V}_k$. In addition, we would like $\boldsymbol{\mu}_{\rho_{k+1}}$ to be closer to the mean of the initial state distribution of the original MDP, $\boldsymbol{\mu}_\rho$. Specifically, we compute $\boldsymbol{\mu}_{\rho_{k+1}}$ by minimizing the following objective function subject to constraints: 
\begin{align}
\label{eqn:discoverNewNode}
\boldsymbol{\mu}_{\rho_{k}} = \argmin_\vc{s} V_{k-1}(\vc{s}) + w \|\vc{s} - \boldsymbol{\mu}_{\rho}\|^2 \nonumber \\
\text{subject to}\;\;\; V_{k-1}(\vc{s}) \geq \bar{V}_{k-1} \nonumber \\
C(\vc{s}) \geq 0 
\end{align}
where $C(\vc{s})$ represents the environment constraints, such as the constraint that enforces collision-free states. Since the value function $V_k$ in Equation \ref{eqn:discoverNewNode} is differentiable through back-propagation, we can use any standard gradient-based algorithms to solve this optimization. Procedure for selecting the weighting coefficient $\bold{w}$ is explained in the appendix.

In addition, we define the objective function of the $(k+1)^{th}$ MDP as follows:
\begin{align}
\label{eqn:relay_obj}
J_{k+1} = \mathbb{E}_{\vc{s}_{0:t_f}, \vc{a}_{0:t_f}}[\sum_{t=0}^{t_f} \gamma^t r(\vc{s}_t, \vc{a}_t) + \textbf{$\alpha$} \gamma^{t_f} g(\vc{s}_{t_f})] ,\hspace{1.5cm} &  \\
\mathrm{where\;\;\;} g(\vc{s}_{t_f}) = \begin{cases} 
V_{k}(\vc{s}_{t_f}), & V_{k}(\vc{s}_{t_f}) > \bar{V}_{k}\\
0 & \mathrm{otherwise}.
\end{cases}. \nonumber
\end{align}

Besides the accumulated reward, this cost function has an additional term to encourage ``relaying''. That is, if the rollout is terminated because it enters the subset of $\mathcal{S}$ where the policy of the parent node is capable of handling (\ie $V_{k}(\vc{s}_{t_f}) > \bar{V}_{k}$), it will receive the accumulated reward by following the parent policy from $\vc{s}_{t_f}$. This quantity is approximated by $V_{k}(\vc{s}_{t_f})$ because it recursively adds the accumulated reward earned by each policy along the chain from $\mathcal{N}_k$ to $\mathcal{N}_0$. If a rollout terminates due to other terminal conditions (\eg falling on the ground for a locomotion task), it will receive no relay reward. Using this objective function, we can learn a policy $\pi_{k+1}$ that drives a rollout towards states deemed good by the parent's value function, as well as a value function $V_{k+1}$ that measures the long-term reward from the current state, following the policies along the chain (Line \ref{line:solve}). Finally, we compute the threshold of the value function (Line \ref{line:threshold}), add a new node, $\mathcal{N}_{k+1}$, to $\Gamma$ (Line \ref{line:addnode}), and add a new edge that connects $\mathcal{N}_{k+1}$ to $\mathcal{N}_k$ (Line \ref{line:addedge}).
 
The weighting parameter $\alpha$ determines the importance of ``relaying'' behavior. If $\alpha$ is set to zero, each MDP will attempt to solve the original task on its own without leveraging previously solved policies. The value of $\alpha$ in all our experiments is set to $30$ and we found that the results are not sensitive to $\alpha$ value, as long as it is sufficiently large (Section \ref{sec:alphaComparison}).

\subsection{Computing Threshold for Value Function}
\label{sec:threshold}

\begin{algorithm}[t]
\caption{ComputeThreshold}\label{alg:computeThreshold}
\begin{algorithmic}[1]
\STATE \textbf{Input:} $\pi, V, \mathcal{T}, \rho = N(\boldsymbol{\mu}_\rho, \Sigma_\rho), J$
\STATE Initialize buffer $\mathcal{D}$ for training data
\STATE $[\vc{s}_1, \cdots, \vc{s}_M] \leftarrow$ Sample states from $N(\boldsymbol{\mu}_\rho, 1.5\Sigma_\rho) $ \label{line:sample}
\STATE $[\tau_1, \cdots, \tau_M] \leftarrow$ Generate rollouts by following $\pi$ and $\mathcal{T}$ from $[\vc{s}_1, \cdots, \vc{s}_M]$ \label{line:generate}
\STATE Compute returns for rollouts: $R_i = J(\tau_i),\; i \in [1,M]$ \label{line:computeReturn}
\STATE $\bar{R} \leftarrow$ Compute average of returns for rollouts not terminated by $\mathcal{T}$ \label{line:averageReturn}
\FOR{$i = 1:M$} \label{line:trainingSetBegin}
\IF{$R_i > \bar{R}$}
\STATE Add $(V(\vc{s}_i), 1)$ in $\mathcal{D}$
\ELSE
\STATE Add $(V(\vc{s}_i), 0)$ in $\mathcal{D}$
\ENDIF 
\ENDFOR \label{line:trainingSetEnd}
\STATE $\bar{V} \leftarrow$ Classify($\mathcal{D}$) \label{line:classify}
\RETURN{$\bar{V}$}
\end{algorithmic}
\end{algorithm}

In practice, $V(\vc{s})$ provided by the learning algorithm is only an approximation of the true value function. We observe that the scores $V(\vc{s})$ assigns to the successful states are relatively higher than the unsuccessful ones, but they are not exactly the same as the true returns. As such, we can use $V(\vc{s})$ as a binary classifier to separate ``good'' states from ``bad'' ones, but not as a reliable predictor of the true returns.

To use $V$ as a binary classifier of the state space, we first need to select a threshold $\bar{V}$. For a given policy, separating successful states from unsuccessful ones can be done as follows. First, we compute the average of true return, $\bar{R}$, from a set of sampled rollouts that do not terminate due to failure conditions. Second, we compare the true return of a given state to $\bar{R}$ to determine whether it is a successful state (successful if the return is above $\bar{R}$). In practice, however, we can only obtain an approximated return of a state via $V(\vc{s})$ during policy learning and execution. Our goal is then to find the optimal $\bar{V}$ such that the separation of approximated returns by $\bar{V}$ is as close as possible to the separation of true returns by $\bar{R}$.


Algorithm \ref{alg:computeThreshold} summarizes the procedure to compute $\bar{V}$. Given a policy $\pi$, an approximated value function $V$, termination conditions $\mathcal{T}$, a Gaussian initial state distribution $\rho = N(\boldsymbol{\mu}_{\rho}, \Sigma_\rho)$, and the objective function of the MDP $J$, we first draw $M$ states from an expanded initial state, $N(\boldsymbol{\mu}_{\rho}, 1.5 \Sigma_\rho)$ (Line \ref{line:sample}), generate rollouts from these sampled states using $\pi$ (Line \ref{line:generate}), and compute the true return of each rollout using $J$ (Line \ref{line:computeReturn}). Because the initial states are drawn from an inflated distribution, we obtain a mixed set of successful and unsuccessful rollouts. We then compute the average return of successful rollouts $\bar{R}$ that do not terminate due to the terminal conditions $\mathcal{T}$ (Line \ref{line:averageReturn}). Next, we generate the training set where each data point is a pair of the predicted value $V(\vc{s}_i)$ and a binary classification label, ``good'' or ``bad'', according to $\bar{R}$ (\ie $R_i > \bar{R}$ means $\vc{s}_i$ is good) (Line \ref{line:trainingSetBegin}-\ref{line:trainingSetEnd} ). We then train a binary classifier represented as a decision tree to find to find $\bar{V}$ (Line \ref{line:classify}).

\subsection{Applying Relay Networks}


Once the graph of relay networks $\Gamma$ is trained, applying the polices is quite straightforward. For a given initial state $\vc{s}$, we select a node $c$ whose $V(\vc{s})$ has the highest value among all nodes. We execute the current policy $\pi_c$ until it reaches a state where the value of the parent node is greater than its threshold ($V_{p(c)}(\vc{s}) > \bar{V}_{p(c)}$), where $p(c)$ indicates the index of the parent node of c. At that point, the parent control policy takes over and the process repeats until we reach the root policy. Alternatively, instead of always switching to the parent policy, we can switch to another policy whose $V(\vc{s})$ has the highest value. 

\subsection{Extending to Multiple Strategies}
\label{sec:multiple-chains}
Optionally, our algorithm can be extended to discover multiple solutions to the original MDP. Take the problem of cartpole swing-up and balance as an example. In Algorithm \ref{alg:relay_networks}, if we choose the mean of $\rho_0$ to be slightly off to the left from the balanced goal position, we will learn a chain of relay networks that often swings to the left and approaches the balanced position from the left. If we run Algorithm \ref{alg:relay_networks} again with the mean of $\rho_0$ leaning toward right, we will end up learning a different chain of polices that tends to swing to the right. For a problem with multi-modal solutions, we extend Algorithm \ref{alg:relay_networks} to solve for a directed graph with multiple chains and describe an automatic method to merge the current chain into an existing one to improve sample efficiency. Specifically, after the current node $\mathcal{N}_k$ is added to $\Gamma$ and the next initial state distribution $\rho_{k+1}$ is proposed (Line \ref{line:next_initial} in Algorithm \ref{alg:relay_networks}), we compare $\rho_{k+1}$ against the initial state distribution stored in every node on the existing chains (excluding the current chain). If there exists a node $\tilde{\mathcal{N}}$ with a similar initial state distribution, we merge the current chain into $\tilde{\mathcal{N}}$ by learning a policy (and a value function) that relays to $\mathcal{N}_k$ from the initial state distribution of $\tilde{\mathcal{N}}$, essentially adding an edge from $\tilde{\mathcal{N}}$ to $\mathcal{N}_k$ and terminating the current chain. Since now $\tilde{\mathcal{N}}$ has two parents, it can choose which of the two policies to execute based on the value function or by chance. Either path will lead to completion of the task, but will do so using different strategies. The details of the algorithm can be found in Appendix A.

\section{Results}

We evaluate our algorithm on motor skill control problems in simulated environments. We use DART physics engine \cite{DART} to create five learning environments similar to Cartpole, Hopper, 2D Walker, and Humanoid environments in Open-AI Gym \cite{OpenAI}. To demonstrate the advantages of relay networks, our tasks are designed to be more difficult than those in Open-AI Gym. Implementation details can be found in the supplementary document. We compare our algorithm to three baselines:
\begin{itemize}[leftmargin=*]
\item{A single policy (ONE)}: ONE is a policy trained to maximize the objective function of the original task from the  initial state distribution $\rho$. For fair comparison, we train ONE with the same number of samples used to train the entire relay networks graph. ONE also has the same number of neurons as the sum of neurons used by all relay policies. 
\item{No relay (NR)}: NR validates the importance of relaying which amounts to the second term of the objective function in Equation \ref{eqn:relay_obj} and the terminal condition, $V_k(\vc{s}) > \bar{V}_k$. NR removes these two treatments, but otherwise identical to relay networks. To ensure fairness, we use the same network architectures and the same amount of training samples as those used for relay networks. Due to the absence of the relay reward and the terminal condition $V_k(\vc{s} > \bar{V}_k$, each policy in NR attempts to achieve the original task on its own without relaying. During execution, we evaluate every value function at the current state and execute the policy that corresponds to the highest value.
\item{Curriculum learning (CL): We compare with curriculum learning which trains a single policy with increasingly more difficult curriculum. We use the initial state distributions computed by Algorithm \ref{alg:relay_networks} to define different curriculum. That is, we train a policy to solve a sequence of MDP's defined as $\{\mathcal{S}, \mathcal{A}, r, \mathcal{T}, \rho_k, P\}$, $k\in[0,K]$, where $K$ is the index of the last node on the chain. When training the next MDP, we use previously solved $\pi$ and $V$ to "warm-start" the learning.}
\end{itemize}

\subsection{Tasks}
We will briefly describe each task in this section. Please see Appendix B in the supplementary document for detailed description of the state space, action space, reward function, termination conditions, and initial state distribution for each problem.

\begin{itemize}[leftmargin=*]
\item{\textbf{Cartpole:}}
Combining the classic cartpole balance problem with the pendulum swing-up problem, this example trains a cartpole to swing up and balance at the upright position. The mean of initial state distribution, $\boldsymbol{\mu}_{\rho}$, is a state in which the pole points straight down and the cart is stationary. Our algorithm learns three relay policies to solve the problem.



\item{\textbf{Hopper:}}
This example trains a 2D one-legged hopper to get up from a supine position and hop forward as fast as possible. We use the same hopper model described in Open AI Gym. The main difference is that $\boldsymbol{\mu}_{\rho}$ is a state in which the hopper lies flat on the ground with zero velocity, making the problem more challenging than the one described in OpenAI Gym. Our algorithm learns three relay policies to solve the problem.



\item{\textbf{2D walker with initial push:}}
The goal of this example is to train a 2D walker to overcome an initial backward push and walk forward as fast as possible. We use the same 2D walker environment from Open AI Gym, but modify $\boldsymbol{\mu}_{\rho}$ to have a negative horizontal velocity. Our algorithm learns two relay policies to solve the problem.

\item{\textbf{2D walker from supine position:}}
We train the same 2D walker to get up from a supine position and walk as fast as it can. $\boldsymbol{\mu}_{\rho}$ is a state in which the walker lies flat on the ground with zero velocity. Our algorithm learns three relay policies to solve the problem.

\item{\textbf{Humanoid walks:}} 
This example differs from the rest in that the subtasks are manually designed and the environment constraints are modified during training. We train a 3D humanoid to walk forward by first training the policy on the sagittal plane and then training in the full 3D space. As a result, the first policy is capable of walking forward while the second policy tries to brings the humanoid back to the sagittal plane when it starts to deviate in the lateral direction. For this example, we allow the policy to switch to non-parent node. This is necessary because while walking forward the humanoid deviates from the sagittal plane many times. 
\end{itemize}

\subsection{Baselines Comparisons}
We compare two versions of our algorithm to the three baselines mentioned above. The first version (AUTO) is exactly the one described in Algorithm \ref{alg:relay_networks}. The second version (MANUAL) requires the user to determine the subtasks and the initial state distributions associated with them. While AUTO presents a cleaner algorithm with less user intervention, MANUAL offers the flexibility to break down the problem in a specific way to incorporate domain knowledge about the problem.

\begin{figure*}[t]
\centering
\setlength{\tabcolsep}{1pt}
\renewcommand{\arraystretch}{0.7}
\begin{tabular}{c c c c}
  \hline
  \includegraphics[width=0.24\textwidth,height=3cm]{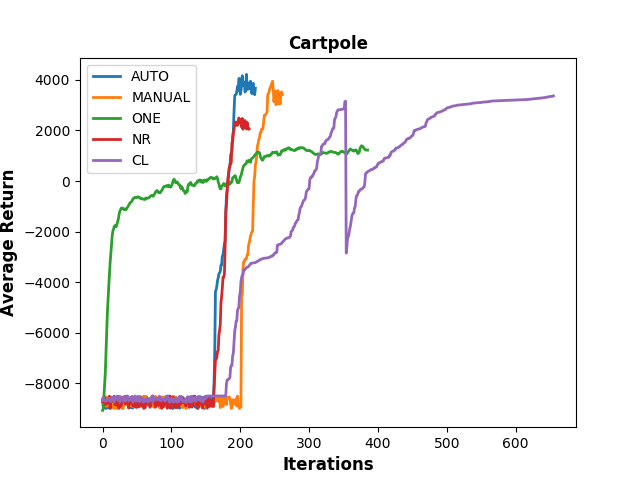}&
  \includegraphics[width=0.24\textwidth,height=3cm]
  {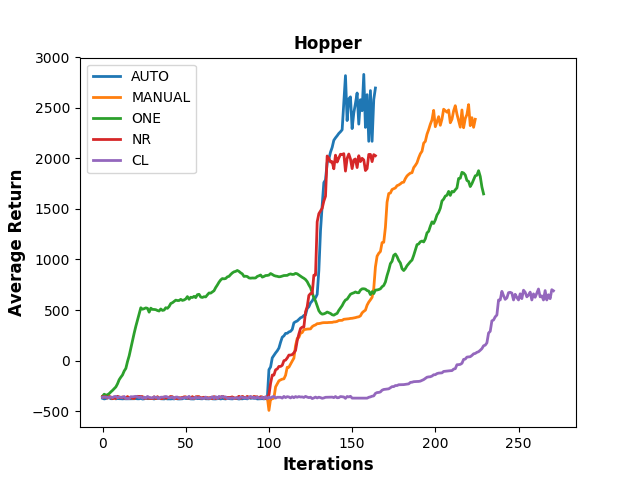}&
  \includegraphics[width=0.24\textwidth,height=3cm]{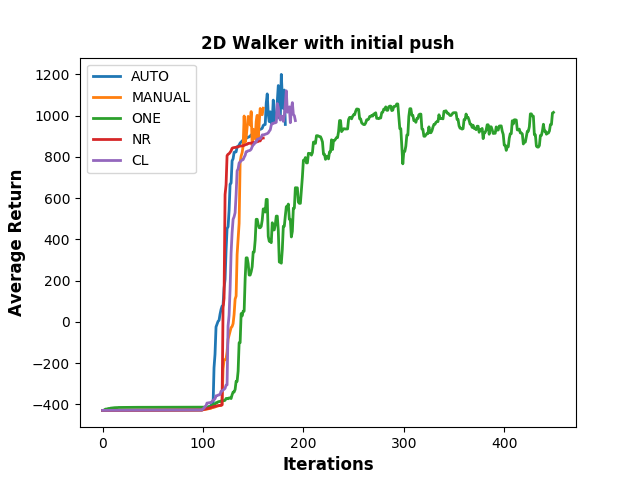}&
  \includegraphics[width=0.24\textwidth,height=3cm]{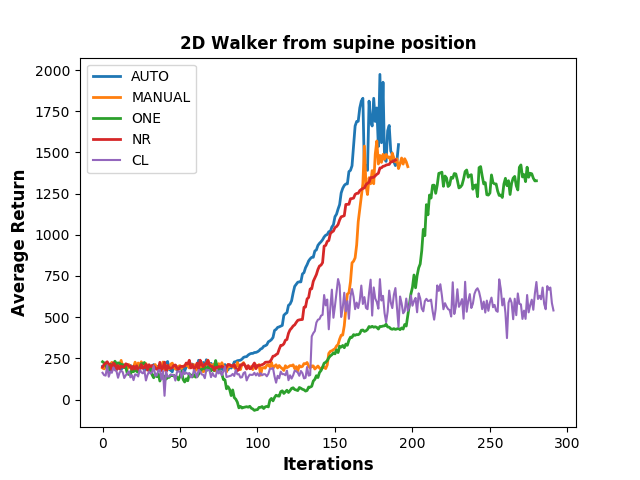} \\ 
  \hline  
\end{tabular}
\caption{ Testing curve comparisons.
    }
\label{fig:TestingCurve}
\end{figure*}

Figure \ref{fig:TestingCurve} shows the \emph{testing curves} during task training. The learning curves are not informative for comparison because the initial state distributions and/or objective functions vary as the training progresses for AUTO, MANUAL, NR, and CL. The testing curves, on the other hand, always computes the average return on the original MDP. That is, the average objective value (Equation \ref{eqn:rootObj}) of rollouts drawn from the original initial state distribution $\rho$. Figure \ref{fig:TestingCurve} indicates that while both AUTO and MANUAL can reach higher returns than  the baselines, AUTO is in general more sample efficient than MANUAL. Further, training the policy to relay to the ``good states'' is important as demonstrated by the comparison between AUTO and NR. The results of CL vary task by task, indicating that relying learning from a warm-started policy is not necessarily helpful.

\subsection{Analyses}
\label{sec:alphaComparison}

\begin{figure*}[t]
\centering
\setlength{\tabcolsep}{1pt}
\renewcommand{\arraystretch}{0.7}
\begin{tabular}{c c c c}
  \hline
  \includegraphics[width=0.26\textwidth,height=3cm]{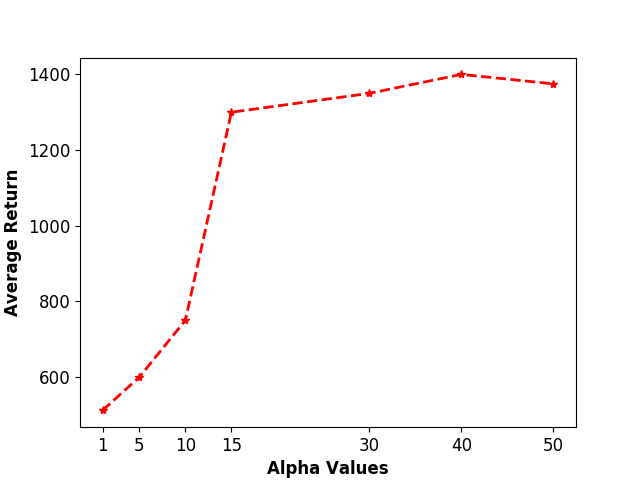}&
  \includegraphics[width=0.26\textwidth,height=3cm]{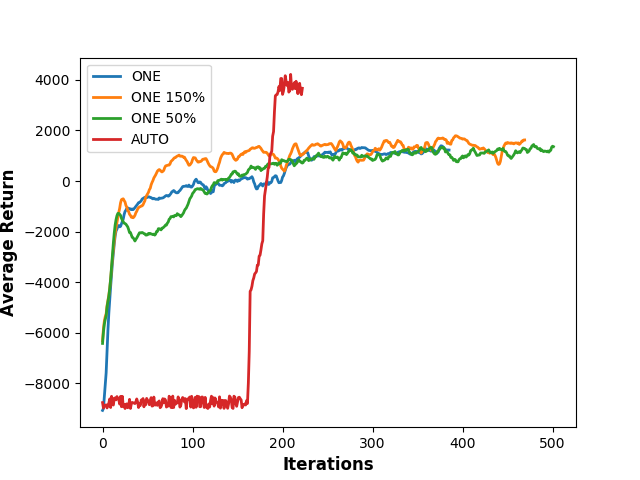}&
  \includegraphics[width=0.2\textwidth]{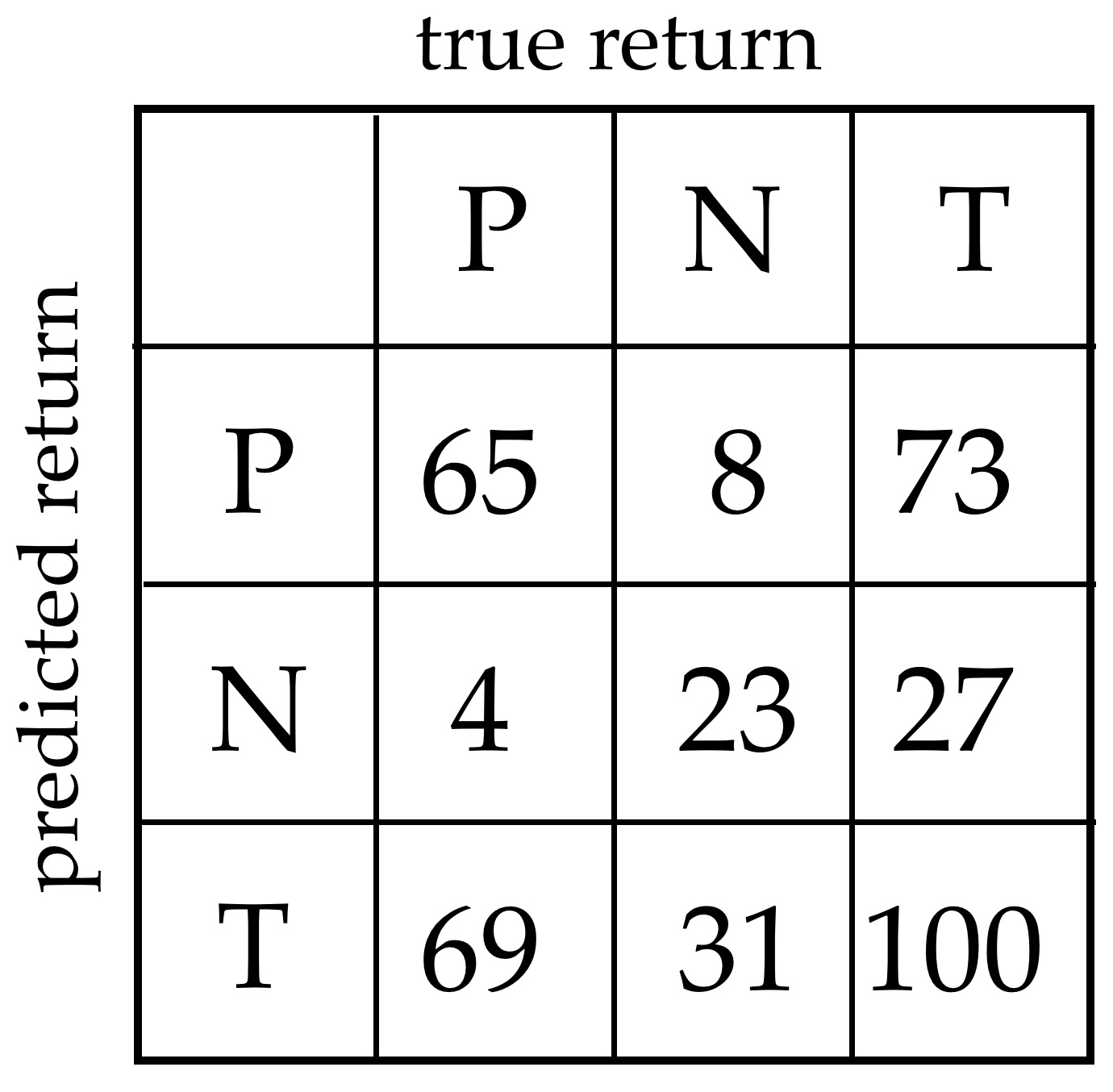}&
  \includegraphics[width=0.2\textwidth]{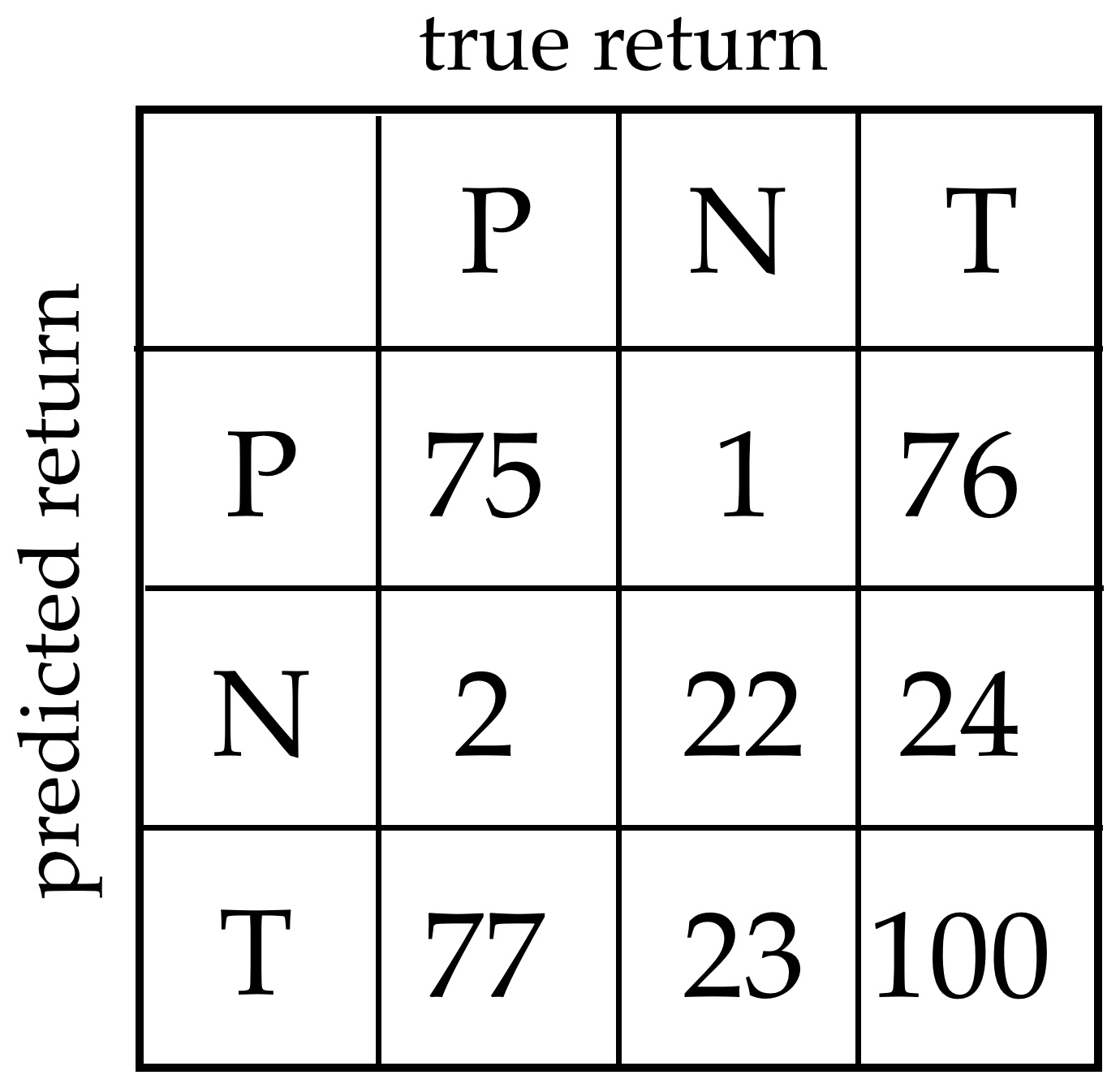} \\
  
 \hline  
\end{tabular}
\caption{ (a) The experiment with $\alpha$. (b) Comparison of ONE with different numbers of neurons. (c) Confusion matrix of value function binary classifier (d) Confusion matrix after additional regression.
    }
\label{OtherFigures}
\end{figure*}

\textbf{Relay reward:} One important parameter in our algorithm is the weight for relay reward, \ie $\alpha$ in Equation \ref{eqn:relay_obj}. Figure \ref{OtherFigures}(a) shows that the policy performance is not sensitive to $\alpha$ as long as it is sufficiently large.

\textbf{Accuracy of value function:} Our algorithm relies on the value function to make accurate binary prediction. To evaluate the accuracy of the value function, we generate $100$ rollouts using a learned policy and label them negative if they are terminated by the termination conditions $\mathcal{T}$. Otherwise, we label them positive. We then predict the rollouts positive if they satisfy the condition, $V(\vc{s}) > \bar{V}$. Otherwise, they are negative. Figure \ref{OtherFigures}(c) shows the confusion matrix of the prediction. In practice, we run an additional regression on the value function after the policy training is completed to further improve the consistency between the value function and the final policy. This additional step can further improve the accuracy of the value function as a binary predictor (Figure \ref{OtherFigures}(d)).

\textbf{Sizes of neural networks:}
The size of the neural network for ONE affects the fairness of our baseline comparison. Choosing a network that is too small can restrict the ability of ONE to solve challenging problems. Choosing a network that is equal or greater in size than the sum of all relay polices might make training too difficult. We compare the performance of ONE with three different network sizes: $N$, $150\%N$, $50\%N$, where $N$ is the number of neurons used for the entire relay networks graph. Figure \ref{OtherFigures}(b) shows that the average return for these three polices are comparable. We thus choose $N$ to be the one for our baseline.

\subsection{Multiple chains}
We test the extended algorithm for multiple chains (Appendix A) on two tasks: Cartpole and Hopper. For the first chain of the Cartpole example, the mean of initial state distribution for the first node is slightly off to the right from the balanced position (Figure \ref{fig:MultipleStratgies} in Appendix A). The algorithm learns the task using three nodes. When training the second chain, we choose the mean of the initial state distribution for the first node to be slightly leaning to the left. The algorithm ends up learning a very different solution to swing up and balance. After learning two subtasks, the second chain identifies that the next subtask is very similar to the existing third subtask on the first chain and thus merges with the first chain. For Hopper, we make the mean of initial state distribution leaning slightly backward for the first chain, and slightly forward for the second chain. The algorithm creates three nodes for the first chain and merges the second chain into the first one after only one policy is learned on the second chain (Figure \ref{fig:MultipleStratgies} in Appendix A). 

To validate our method for automatic merging nodes, we also learn a graph that does not merge nodes ($\Gamma'$). The average return of taking the second chain after the junction in $\Gamma$ is comparable to the average return of the second chain in $\Gamma'$. However, $\Gamma$ requires less training samples than $\Gamma'$, especially when the merge happens closer to the root. In the two examples we test, merging saves 4K training samples for Cartpole and 50K for Hopper.

\section{DICUSSION}
We propose a technique to learn a robust policy capable of controlling a wide range of state space by breaking down a complex task to simpler subtasks. Our algorithm has a few limitations. The value function is approximated based on the visited states during training. For a state that is far away from the visited states, the value function can be very inaccurate. Thus, the initial state distribution of the child node cannot be too far from the parent's initial state distribution. In addition, as mentioned in the introduction, the relay networks are built on locally optimal policies, resulting globally suboptimal solutions to the original task. The theoretical bounds of the optimality of relay networks can be an interesting future direction.



\clearpage
\acknowledgments{We want to thank the reviewers for their feedback. This work was supported by NSF award IIS-1514258, Samsung Global Research Outreach grant and AWS Cloud Credits for Research.}


\bibliography{references}  

\clearpage
\section*{Appendix A: Multiple Strategies Algorithm}

\begin{figure*}[!htb]
\centering
\includegraphics[width=0.5\textwidth]{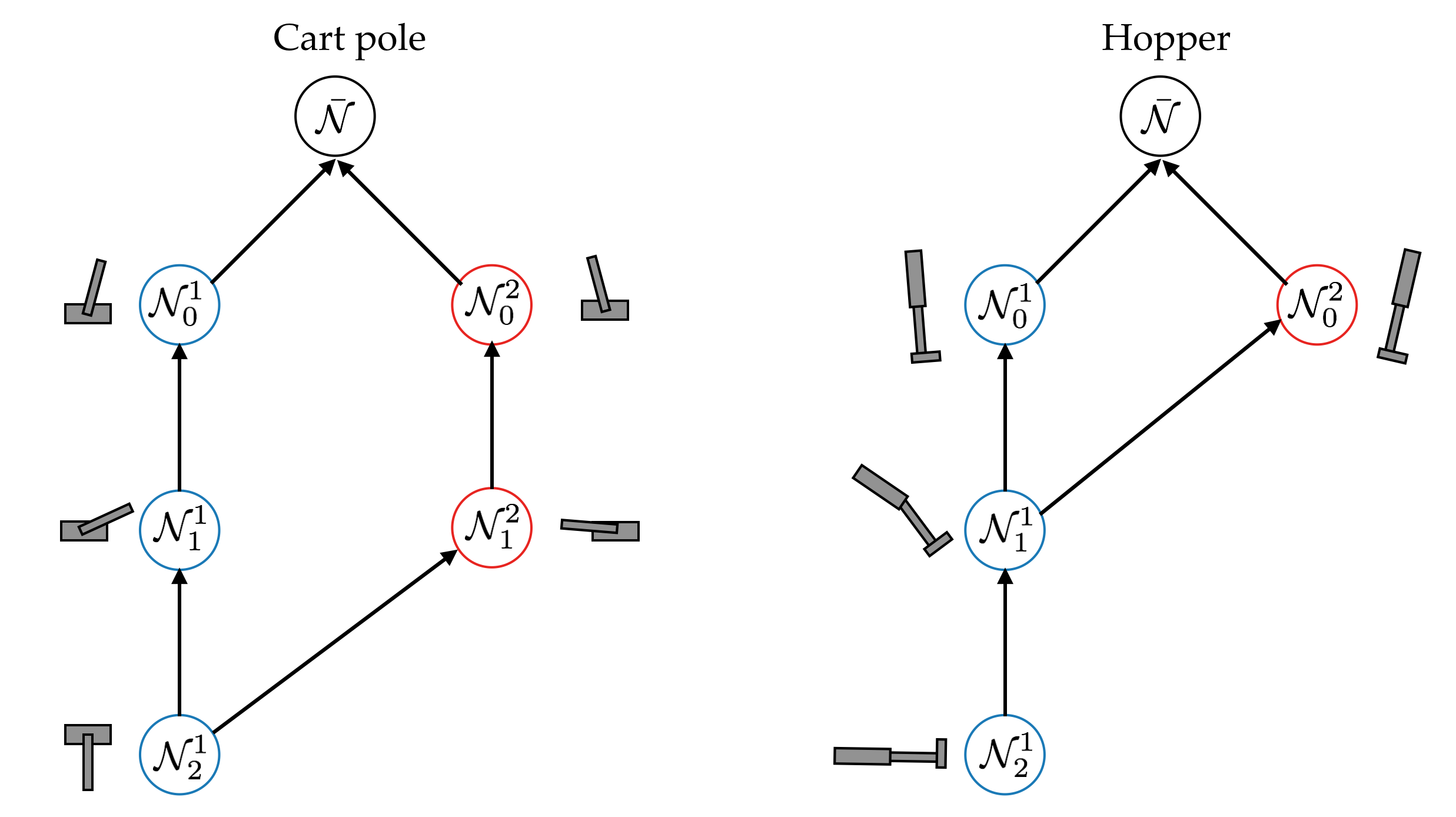}
\caption{ The generated directed graph for the Cartpole (Left) and the Hopper (Right) tasks. Each small image illustrates the mean of the initial state distribution of the corresponding node.
} 
\label{fig:MultipleStratgies}
\end{figure*}

This appendix provides additional information on finding multiple solutions to the original MDP (\secref{multiple-chains}).
\algref{multi_relay_networks} is the pseudo-code of the described algorithm and \figref{MultipleStratgies} shows the directed graphs generated  for the Cartpole and the Hopper tasks.

\begin{algorithm}[!htb]
\caption{LearnMultiRelayNetworks}\label{alg:multi_relay_networks}
\begin{algorithmic}[1]
\STATE \textbf{Input:} MDP $\{\mathcal{S}, \mathcal{A}, r, \mathcal{T}, \rho, P\}$
\STATE $\Gamma = \emptyset, \mathcal{D} = \emptyset$
\STATE Add root node, $\bar{\mathcal{N}} = \{\emptyset\}$, to $\Gamma$
\FOR {$i=1:nSolutions$}
\STATE Define a simpler initial state distribution $\rho_0$ 
\STATE Define objective function $J_0$ according to Equation \ref{eqn:rootObj}
\STATE $[\pi_0, V_0] \leftarrow$ \textrm{PolicySearch}($\mathcal{S}, \mathcal{A}, \rho_0, J_0, \mathcal{T})$
\STATE $\bar{V}_0 \leftarrow$
ComputeThreshold($\pi_0, V_0, \mathcal{T}, \rho_0, J_0$)
\STATE Add node, $\mathcal{N}^i_0 = \{\rho_0\}$, to $\Gamma$
\STATE Add edge, $\mathcal{E} = \{\pi_0, V_0, \bar{V}_0, \}$, from $\mathcal{N}^i_0$ to $\bar{\mathcal{N}}$
\STATE $k=0$
\WHILE{$\pi_k$ does not succeed from $\rho$}
\STATE $\mathcal{T}_{k+1} = \mathcal{T} \cup (V_{k}(\vc{s}) > \bar{V}_{k})$
\STATE $\rho_{k+1} \leftarrow$ Compute new initial state distribution using Equation \ref{eqn:discoverNewNode}
\STATE Define objective function $J_{k+1}$ according to Equation \ref{eqn:relay_obj}
\IF {there exists $\tilde{\mathcal{N}} \in \mathcal{D}$ such that $D_{KL}(\tilde{\mathcal{N}}.\boldsymbol{\mu}, \boldsymbol{\mu}_{\rho_{k+1}}) < \epsilon$ }
\STATE $[\pi_{k+1}, V_{k+1}] \leftarrow$ \textrm{PolicySearch}($\mathcal{S}, \mathcal{A}, \tilde{\mathcal{N}}.\boldsymbol{\mu}, J_{k+1}, \mathcal{T}_{k+1})$
\STATE $\bar{V}_{k+1} \leftarrow$
ComputeThreshold($\pi_{k+1}, V_{k+1}, \mathcal{T}, \tilde{\mathcal{N}}.\boldsymbol{\mu}, J_{k+1}$)
\STATE Add edge $\mathcal{E} = \{\pi_{k+1}, V_{k+1}, \bar{V}_{k+1}\}$ from node $\tilde{\mathcal{N}}$ to $\mathcal{N}^i_k$
\STATE \textbf{break}
\ENDIF
\STATE $[\pi_{k+1}, V_{k+1}] \leftarrow$ \textrm{PolicySearch}($\mathcal{S}, \mathcal{A}, \rho_{k+1}, J_{k+1}, \mathcal{T}_{k+1})$
\STATE $\bar{V}_{k+1} \leftarrow$
ComputeThreshold($\pi_{k+1}, V_{k+1}, \mathcal{T}, \rho_{k+1}, J_{k+1}$)
\STATE Add node, $\mathcal{N}^i_{k+1} = \{\rho_{k+1}\}$, to $\Gamma$
\STATE Add edge, $\mathcal{E} = \{\pi_{k+1}, V_{k+1}, \bar{V}_{k+1}\}$, from node $\mathcal{N}^i_{k+1}$ to $\mathcal{N}^i_k$
\STATE $k \leftarrow k+1$
\ENDWHILE
\STATE Add $[\mathcal{N}^i_0, \cdots, \mathcal{N}^i_k]$ to $\mathcal{D}$
\ENDFOR
\RETURN{$\Gamma$}
\end{algorithmic}
\end{algorithm}

\section*{Appendix B: Task Descriptions}

\subsection*{B.1 Cartpole}
This example combines the classic cartpole balance problem with the pendulum swing-up problem. 
Our goal is to train a cartpole to be able to swing up and balance by applying only horizontal forces to the cart. 
The state space of the problem is $[\theta,\dot{\theta},x,\dot{x}]$, where $\theta$ and $\dot{\theta}$ are the angle and velocity of the pendulum and $x$ and $\dot{x}$ are the position and velocity of the the cart. 
The reward function of the problem is to keep the pole upright and the cart close to the origin:
\begin{equation}
r_i =  cos(\theta)  - x^2 - 0.01\|\boldsymbol{\tau}\|^{2} + \kappa.
\end{equation}

Here $\kappa$ ($=2.0$) is an alive bonus and $\boldsymbol{\tau}$ is the control cost. The termination conditions $\mathcal{T}$ in the MDP's before augmented by Algorithm \ref{alg:relay_networks} include the timeout condition at the end of rollout horizon $T_{max}$. For the root note, $\mathcal{T}_0$ also includes the condition that terminates the rollout if the pendulum falls below the horizontal plane. 

\subsection*{B.2 Hopper}
In this problem, our goal is to train a 2D one-legged hopper to get up from a supine position and hop forward as fast as possible.
The hopper has a floating base with three unactuated DOFs and a single leg with hip, knee, and ankle joints.
The 11D state vector consists of $[y, \theta, \vc{q}_{leg}, \dot{x}, \dot{y}, \dot{\theta}, \vc{\dot{q}}_{leg}]$, where $x$, $y$, and $\theta$ are the torso position and orientation in the global frame.
The action is a vector of torques $\boldsymbol{\tau}$ for leg joints.
Note that the policy does not dependent on the global horizontal position of the hopper.
The reward function is defined as:
\begin{equation}
	r_i = \dot{x} - 0.001\|\boldsymbol{\tau}\|^{2} + \frac{1}{1 + |\theta|} + \kappa.
    \label{eq:Hop}
\end{equation} 

Equation \ref{eq:Hop} encourages fast horizontal velocity, low torque usage and maintaining an upright orientation.
The termination conditions for the root policy $\mathcal{T}_0$ include the timeout condition and the failing condition that terminates the hopper when it leans beyond a certain angle in either the forward or backward direction.

\subsection*{B.3 2D walker under initial push}
This task involves a 2D walker moving forward as fast as possible without losing its balance.
The 2D walker has two legs with three DOFs in each leg.
Similar to the previous problem, the 17D state vector includes $[y, \theta, \vc{q}_{leg}, \dot{x}, \dot{y}, \dot{\theta}, \vc{\dot{q}}_{leg}]$ without the global horizontal position $x$. The action is the joint torques $\boldsymbol{\tau}$.
The reward function encourages the positive horizontal velocity, penalizes excessive joint torques and also encourages maintaining an upright orientation. :
\begin{equation}
	r_i = \dot{x} - 0.001\|\boldsymbol{\tau}\|^{2} +  \frac{1}{1 + |\theta|} +  \kappa.
	\label{eq:reward_2d}
\end{equation}
While training the root policy, we terminate the rollout when the orientation of the 2D-walker exceeds a certain angle or if the timeout condition is met. 

\subsection*{B.4 2D walker from supine position}
This tasks uses the same 2D Walker robot with the same state space as described in the previous section. 
It also uses the same reward function described in Equation \ref{eq:reward_2d}. 
The termination condition for the root policy $\mathcal{T}_{0}$ is met if either the rollout satisfies the timeout condition or if the robot exceeds a certain orientation. 

\subsection*{B.5 Humanoid walks}
The 3D Walker problem has the similar goal of moving forward without falling.
The robot has a total of $21$ DOFs with six unactuated DOFs for the floating base, three DOFs for the torso, and six DOFs for each leg. Therefore, the 41D state vector includes $[y, z, \vc{r}, \vc{q}_{torso}, \vc{q}_{leg}, \dot{x}, \dot{y}, \dot{z}, \vc{\dot{r}}, \vc{\dot{q}}_{torso}, \vc{\dot{q}}_{leg}]$ where $x$, $y$, $z$ are the global position and $\vc{r}$ is the global orientation of the robot. The action is the joint torques $\boldsymbol{\tau}$.
The reward function for this problem is:
\begin{equation}
	r_i = \dot{x} - 0.001\|\boldsymbol{\tau}\|^{2}\  - 0.2|c_{y}| + \kappa \;, \mathrm{where\;} i = \{0,1\},
    \label{eq:3dbiped}
\end{equation}
where $c_y$ is the deviation of center of mass. 
While training the root policy, we terminate the rollout when the orientation of the walker exceeds a certain angle or if the timeout condition is met.

\section*{Appendix C: Implementation Details}
\begin{figure*}[!htb]
\centering
\includegraphics[width=0.5\textwidth]{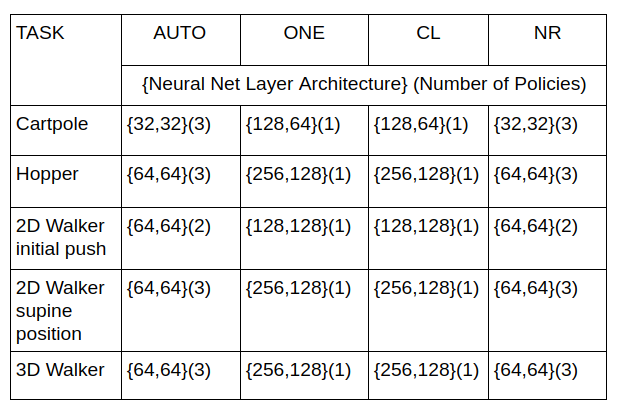}
\caption{ Different neural network architectures used in the training for each task.
} 
\label{fig:NetSize}
\end{figure*}

We use DartEnv \cite{dartenv}, which is an Open-AI environment that uses PyDart \cite{pydart} as a physics simulator. PyDart is an open source python binding for Dynamic Animation and Robotics Toolkit (DART) \cite{DART}. The time step of all the simulations are 0.002s. To train the policy, we use \emph{baselines} \cite{OpenAIBaselines} implementation of PPO. We provide the hyperparameters of neural networks in Figure \ref{fig:NetSize}.

\section*{Appendix D: Weighting Coefficient}
As described in section \ref{sec:newnodes}, we generate new nodes in the relay by defining an optimization problem. The objective function of this is described in Equation \ref{eqn:discoverNewNode}. Just like many optimization problems, the weights of the objective terms are often determined empirically. We determine the value of $\textit{w}$ in Equation \ref{eqn:discoverNewNode} based on the following procedure. For example, after the root policy $\pi_{0}$ is trained, we look at the magnitude of the value function $V_0(\vc{s})$ for a few successful states sampled near $\rho_{0}$. Then the value of $\textit{w}$ is set such that the magnitude of the two terms $V_0(\vc{s})$ and $\|\vc{s} - \boldsymbol{\mu}_{\rho}\|^2$ of the objective function are comparable. In our experiments, we found that training the relay networks is not sensitive to the value of $\textit{w}$ across sub-tasks. However, the values can be different for different tasks (e.g. hopping vs walking).

\end{document}